\documentclass{article}

\usepackage[final,sglblindworkshop]{neurips_2026}
\workshoptitle{Who Verifies the Agents? Toward Reliable Agent Development}
\usepackage[utf8]{inputenc}
\usepackage[T1]{fontenc}
\usepackage{hyperref}
\usepackage{url}
\usepackage{booktabs}
\usepackage{amsfonts}
\usepackage{amsmath}
\usepackage{microtype}
\usepackage{xcolor}
\usepackage{graphicx}
\usepackage{float}

\title{Auditing Web Agent Evaluation on WebArena-Lite: Human Review of Outcomes and Trajectories}
\author{%
  \parbox{\dimexpr\textwidth-2\tabcolsep\relax}{\centering
    Chengguang Gan$^{1}$ \quad Zimeng He$^{2}$ \quad Yoshihiro Tsujii$^{1}$\\[4pt]
    Ken-ichiro Kobayashi$^{1}$ \quad Hiroki Itoh$^{1}$ \quad Kotaro Funakoshi$^{2}$\\[6pt]
    \normalfont $^{1}$Techtouch, Inc.\\
    $^{2}$Institute of Science Tokyo
  }%
}

\date{}

\hypersetup{
  hidelinks,
  pdftitle={Auditing Web Agent Evaluation on WebArena-Lite: Human Review of Outcomes and Trajectories},
  pdfauthor={Chengguang Gan, Zimeng He, Yoshihiro Tsujii, Ken-ichiro Kobayashi, Hiroki Itoh, Kotaro Funakoshi},
  pdfsubject={Accepted poster at the NeurIPS 2026 Verify-Agents workshop},
  pdfkeywords={web agents, agent evaluation, human-grounded verification, WebArena-Lite}
}

\begin{document}
\maketitle

% Approximately 0.4 page
\begin{abstract}
Web agents are an important application of large language models, yet their evaluation often depends on rule based or language model evaluators that inspect only the final outcome. Human verification of task completion and detailed analysis of failed trajectories remain limited. We audit all 165 WebArena Lite tasks under six evaluation conditions built from GPT 5.5 and an untrained Qwen3.5 9B model. The audit retains the original score, corrects false negatives from the automatic evaluator, identifies the first consequential error, and examines progress across the trajectory. We also study a Memory and Analysis Support Mechanism (MASM), which maintains explicit execution state, and Guide Text, which provides task relevant procedural guidance. Across four GPT 5.5 settings, human review recovers 5.45 to 8.49 percentage points of success missed by the evaluator. With a 25 step budget, Guide Text raises corrected success with MASM from 34.55\% to 38.18\%. On the untrained Qwen3.5 9B model, MASM raises the evaluator score from 13.90\% to 18.80\%. Review of 102 failed GPT 5.5 trajectories reveals frequent scrolling loops, unfinished exploration, premature answers, invalid actions, and incomplete form workflows. Step level evidence further shows that substantial early progress can coexist with a final failure. These results show why final scores alone provide an incomplete account of web agent behavior and motivate human grounded, trajectory aware verification.
\end{abstract}

% Approximately 0.8 page
\section{Introduction}
Large language model agents are increasingly expected to interpret goals, gather information, choose actions, and respond to changes in an external environment. Web agents provide a concrete setting for studying these abilities because many everyday and professional activities are mediated through browser interfaces. Early work explored web interaction from demonstrations and language instructions, while later systems combined reasoning and acting to support longer sequences of decisions \citep{yao2022react,deng2023mind2web}. WebArena has become a central interactive benchmark: it provides reproducible websites for shopping, social discussion, software development, content management, and map based search, together with tasks whose completion can be checked against the environment state \citep{zhou2024webarena}. WebArena Lite, introduced as part of VisualAgentBench \citep{liu2025visualagentbench}, retains 165 representative tasks and makes repeated evaluation more practical.

Progress is usually summarized by the proportion of tasks accepted by an automatic evaluator. This number is useful, but it combines two distinct capabilities. One is whether the agent actually completes the task. The other is whether the evaluator recognizes that completion. Exact string matching, output formatting, partial state checks, and differences in stopping behavior can all change the recorded score without changing what the agent achieved. Unified environments such as BrowserGym improve execution consistency \citep{chezelles2024browsergym}, but they do not guarantee that every final judgment is correct.

This distinction raises three questions. First, how often does the evaluator reject a run that satisfies the user request? Second, when a task truly fails, where does the failure begin and what behavior follows from it? Third, how much verified progress is made before the final outcome is assigned? A binary score cannot distinguish an agent that never reaches the relevant page from one that completes most requirements, reaches a correct state and then continues acting, or returns a semantically correct answer in an incompatible format.

We study these questions through a human grounded audit of all 165 WebArena Lite tasks. We retain the original evaluator result and review failed runs against the task instruction, environment state, screenshots, execution records, and final response. We identify the first consequential error and assign a fine grained failure category. We also decompose tasks into verifiable subgoals and inspect progress after each action. The retained experiments are evaluation only: they compare GPT 5.5 under matched MASM and Guide Text conditions at 15 and 25 steps, together with an untrained Qwen3.5 9B ablation for MASM. No GPT generated trajectory is used to train or improve another model in the experiments reported here.

The audit reveals a consistent gap between measured and observed performance. Human review recovers 5.45 to 8.49 percentage points across the four GPT 5.5 settings. Under a matched 25 step budget, Guide Text raises corrected success with MASM from 57 of 165 tasks (34.55\%) to 63 of 165 (38.18\%), although paired task flips show both gains and regressions. Among 102 failed GPT 5.5 runs in the strongest audited setting, scrolling loops, unfinished exploration, premature answers, invalid action formats, and incomplete workflows dominate. Progress records show that the first actions often produce the largest gains, whereas later steps are more often spent on verification, recovery, and repeated exploration. These findings are intended to improve how web agents are evaluated and debugged, rather than to make a leaderboard claim.

Our contributions are:
\begin{itemize}
    \item We conduct a human grounded audit of all 165 WebArena Lite tasks under six evaluation conditions and quantify false negatives in four matched GPT 5.5 settings.
    \item We identify the first consequential error in failed runs and provide a fine grained taxonomy that separates navigation, information gathering, action execution, stopping, and answer generation failures.
    \item We assess task progress at the step level and show that the same final binary score can hide markedly different trajectories.
    \item We use paired task evidence to analyze MASM, Guide Text, and action budget, and derive practical recommendations for evaluator design, stopping criteria, and deployment monitoring.
\end{itemize}

\section{Related Work}
Web agent research has progressed from offline demonstrations to interactive environments. Mind2Web and WebLINX collect real web interactions, while AgentBench, WebArena, VisualWebArena, WorkArena, and Online Mind2Web test agents on executable or live websites \citep{deng2023mind2web,lu2024weblinx,liu2024agentbench,zhou2024webarena,koh2024visualwebarena,drouin2024workarena,xue2025illusion}. WebArena Lite selects 165 representative tasks for a more practical evaluation cycle, and BrowserGym provides shared interfaces for controlled experiments \citep{liu2025visualagentbench,chezelles2024browsergym}. These resources support comparison across agents, but most reported results still reduce each trajectory to a final success value.

Training methods have advanced from prompted reasoning to learning from browser experience. ReAct alternates reasoning and actions, NNetNav generates additional interaction data, and recent methods such as WebRL and WebAgent R1 apply reinforcement learning to long browser tasks \citep{yao2022react,murty2025nnetnav,qi2025webrl,wei2025webagent}. Other work uses fine grained constraints to select partially successful trajectories for training \citep{logeswaran2026scaling}. These methods improve browser policies, yet their gains are commonly measured with the same automatic signals used during development.

Web benchmarks usually check final answers, URLs, database fields, or page states. LLM judges offer greater flexibility, but previous studies report position effects, self preference, and inconsistent judgments \citep{zheng2023judging,wang2024large}. AgentRewardBench confirms this problem for web agents and finds that rule based evaluators can underreport successful completion \citep{lu2025agentrewardbench}. AgentBoard measures partial progress, while WebGraphEval analyzes recurring paths and critical decisions across trajectories \citep{ma2024agentboard,qian2025webgrapheval}. Together, these studies motivate evaluation that examines both outcomes and the process that produced them.

Our work conducts a human grounded audit of the complete WebArena Lite task set rather than proposing another leaderboard method. We retain both original and human corrected outcomes, identify the first consequential error, categorize failure patterns, and estimate progress at each step. We use matched evaluation conditions to study MASM, Guide Text, and action budget. This analysis separates changes in agent behavior from evaluator artifacts and turns failed trajectories into practical guidance for evaluator design, stopping criteria, and deployment.

\section{Study Scope and Experimental Design}
\label{sec:design}

We study evaluation reliability and trajectory behavior on the fixed set of 165 WebArena Lite tasks. Every condition is aligned by task identifier, which permits paired analysis of gains, regressions, action counts, and stopping behavior. The retained study varies prompt support and action budget without updating model weights. This keeps the scope centered on direct API evaluation and on the reliability of the resulting judgments.

\subsection{WebArena-Lite Evaluation Setting}

WebArena Lite contains 165 tasks drawn from shopping, shopping administration, Reddit, GitLab, and map environments. We use the same task definitions and identifiers in every comparison. Before evaluation, the website containers are reset and the corresponding authentication states are recreated. A run receives a maximum budget of either 15 or 25 actions and may terminate earlier when the agent returns a final answer or explicitly stops.

Table~\ref{tab:main-results} reports six evaluation conditions. Four use GPT 5.5: MASM alone and Guide Text with MASM, each under 15 and 25 step budgets. Two additional conditions use an untrained Qwen3.5 9B model with and without MASM at 25 steps, providing a direct ablation of structured memory support without behavior cloning or parameter updates. For every step, we retain the task instruction, current URL, simplified page representation, selected action, execution status, before and after screenshots, and final response. MASM conditions additionally retain the action analysis and updated memory.

The original WebArena evaluator is run first. Failed GPT 5.5 trajectories are then reviewed under the protocol in Section~\ref{sec:verification}. We report both the original and human corrected outcomes for these four settings. The Qwen ablation is used only to measure the effect of MASM under the original evaluator. All experiments reported in this version evaluate the originating model directly; no recorded GPT output is used to train, fine tune, distill, or otherwise improve another model.

\subsection{MASM and Guide Text}

MASM adds two structured operations after each browser action. First, an action analysis module compares the observed transition with the agent's intention. Its inputs include the task, executed action, execution status or error, before and after URLs, visible page evidence, and screenshots. It returns a structured record with the intended effect, observed result, a judgment of \texttt{yes}, \texttt{partial}, or \texttt{no}, and supporting notes. Second, a memory update module combines the current memory, the most recent five action records, and the latest analysis. It records verified information, feedback on the last step, estimated progress, next steps, and obstacles. The resulting memory is supplied to the next action prompt.

Guide Text is static workflow evidence derived from prior task procedures. It describes useful navigation routes, required fields, and verification checkpoints, but it does not contain the target answer. As shown in Figure~\ref{fig:masm-prompt}, Guide Text enters action planning and memory update. It does not directly enter action result analysis. This separation matters because the analysis should judge the observed transition rather than repeat the suggested procedure. The action model still receives the current page state, action history, previous action, and memory, and must emit one executable browser action. We compare MASM alone with Guide Text and MASM under matched models and action budgets.

\begin{figure*}[t]
    \centering
    \setlength{\fboxsep}{6pt}
\small
\begin{tabular}{c@{\hspace{0.7em}$\longrightarrow$\hspace{0.7em}}c}
\fbox{\begin{minipage}[t]{0.40\textwidth}
\raggedright
\textbf{1. Step inputs}\\[3pt]
\textbf{Task:} What are the top three best selling products in Jan. 2023?\\
\textbf{Guide Text:} Open Reports $\rightarrow$ Bestsellers; set Period to Month; enter 01/01/23--01/31/23; run the report and verify three rows.\\
\textbf{Current state:} URL, simplified HTML, action history, previous action, and current memory.
\end{minipage}} &
\fbox{\begin{minipage}[t]{0.40\textwidth}
\raggedright
\textbf{2. Action prompt and model output}\\[3pt]
\texttt{Task Instruction: \{task\}}\\
\texttt{Action History: \{history\}}\\
\texttt{Agent Memory: \{memory\}}\\
\texttt{Simplified HTML: \{page\_text\}}\\
\texttt{HINT / workflow data: \{guide\_text\}}\\[2pt]
\textbf{Output:} \texttt{do(action="Click", element="549")}
\end{minipage}}
\end{tabular}

\vspace{0.7em}
\begin{tabular}{c@{\hspace{0.45em}$\longrightarrow$\hspace{0.45em}}c@{\hspace{0.45em}$\longrightarrow$\hspace{0.45em}}c}
\fbox{\begin{minipage}[t]{0.245\textwidth}
\raggedright
\textbf{3. Execute and observe}\\[3pt]
Run the selected action and record before and after screenshots, URLs, visible page evidence, execution status, and errors.
\end{minipage}} &
\fbox{\begin{minipage}[t]{0.245\textwidth}
\raggedright
\textbf{4. Action result analysis}\\[3pt]
Inputs: task, action, status, and observed transition.\\
Output: intended effect, observed result, and \texttt{yes}, \texttt{partial}, or \texttt{no}.
\end{minipage}} &
\fbox{\begin{minipage}[t]{0.245\textwidth}
\raggedright
\textbf{5. Memory update}\\[3pt]
Inputs: recent actions, latest analysis, current memory, and Guide Text.\\
Output: verified facts, progress, next steps, and obstacles for the next action prompt.
\end{minipage}}
\end{tabular}

\vspace{0.4em}
\textit{Guide Text supports action planning and memory update, but action result analysis is grounded only in the executed transition and environment evidence.}
    \caption{Prompt and state flow for Guide Text and MASM. Guide Text supports action planning and memory update, while action result analysis is grounded in the executed transition and environment evidence.}
    \label{fig:masm-prompt}
\end{figure*}

\section{Human-Grounded Verification Protocol}
\label{sec:verification}

Automatic success is only the first stage of our evaluation. We retain the original score, then audit the final outcome and the trajectory that produced it. The protocol has three linked parts: correction of evaluator false negatives, annotation of the first consequential trajectory error, and estimation of verified progress after every action. The same evidence hierarchy is used throughout so that outcome correction and process analysis remain consistent.

\subsection{Automated Evaluation and Human Correction}

The WebArena evaluator first assigns the original binary score. We then review every failed task and model pair. A failure is changed to success only when the task requirements are demonstrably satisfied but the evaluator rejects the run because of answer formatting, an overly strict string match, or another documented evaluator limitation. Semantically incomplete answers, unexecuted intentions, and claims supported only by the agent's own memory are not corrected.

Reviewers follow an evidence hierarchy. Environment state and task specific records are preferred when available, followed by before and after screenshots, action execution results, the final answer, and the model's Memory and Analysis. The last source is auxiliary evidence and never overrides the observed environment. Each correction records the applicable rule and the supporting evidence. We preserve both raw and corrected scores, which exposes evaluator sensitivity without replacing the benchmark result. The complete correction rules and task outcome matrix are provided in the appendix.

\subsection{Trajectory Review and Failure Annotation}

For failed runs, reviewers inspect the ordered action list, before and after screenshots, URL changes, action analyses, memory updates, and the final response in a side by side review interface. They identify the \emph{first consequential error}: the earliest action, omission, or unsupported belief that materially reduces the chance of completing the task within the remaining action budget. Harmless exploratory steps are not labeled as errors merely because they do not immediately increase progress.

Each failed task and model pair receives one primary failure mode and a free text note for secondary effects. The taxonomy contains 16 fine grained categories. Major examples include scroll oscillation, unfinished exploration, premature answering, selection of a wrong value or entity, clicks that do not change the page state, exact action repetition, repeated search queries, incomplete form workflows, invalid action parameters, navigation loops, unsupported guesses, and strict answer format mismatches. Reviewers also mark whether the task became complete before the agent stopped. This separates failures of task execution from failures of stopping and final answer production.

Ambiguous cases are flagged for a second review. Agreement statistics are computed only on the subset that is independently double annotated; remaining disagreements are resolved by inspecting the same recorded evidence. This design keeps the taxonomy reproducible while avoiding a stronger agreement claim than the available annotations support.

\subsection{Step-Level Progress Assessment}

Binary success does not distinguish a trajectory that makes no useful progress from one that completes most requirements or finishes the task before stopping. We therefore decompose each task into a small set of verifiable subgoals derived from the instruction and the task specific evaluator. Let task $i$ contain $m_i$ subgoals, and let $z_{i,k,t}\in\{0,1\}$ indicate whether subgoal $k$ is verified after step $t$. We define
\begin{equation}
P_i(t)=\frac{1}{m_i}\sum_{k=1}^{m_i}z_{i,k,t},
\qquad
\Delta_i(t)=P_i(t)-P_i(t-1).
\end{equation}
Unless the task exposes an explicit weighted condition, subgoals receive equal weight. A subgoal is counted only when it is supported by the recorded environment transition. Missing or ambiguous evidence is treated as incomplete.

A positive $\Delta_i(t)$ indicates verified progress, while a negative value records regression caused by leaving a correct state, overwriting a valid input, or unnecessary backtracking. We define the completion step $\tau_i$ as the first step for which all required subgoals are verified. When $P_i(\tau_i)=1$ but the run later receives a failure, the reviewer distinguishes an evaluator false negative from a completion before stopping error by checking whether the completed state persists and whether the required final response is produced.

Memory and Analysis are used to recover the agent's intended subgoal and to locate candidate evidence, but they are not treated as ground truth. Progress decisions are grounded in task specific rubrics, environment state, screenshots, and action effects. This step level view supports comparisons of exploration efficiency, regressions, and stopping behavior even when two runs receive the same final binary score.

% Approximately 3.6 pages
\section{Results and Analysis}
\label{sec:results}

The results address two related questions. The first is how MASM, Guide Text, and action budget change measured success. The second is how much of that measurement depends on the evaluator, and what the trajectories reveal when the final judgment is failure.

\subsection{Overall Performance and Human-Corrected Success}

Table~\ref{tab:main-results} reports all retained evaluation conditions as counts and percentages. GPT 5.5 with Guide Text and MASM at 25 steps gives the highest audited result: the original evaluator accepts 51 of 165 tasks (30.91\%), while human review verifies 63 (38.18\%). The corresponding MASM only run rises from 43 (26.06\%) to 57 (34.55\%) after review.

\begin{table*}[t]
\centering
\caption{Retained WebArena Lite evaluation results. Raw and Corr. report successful tasks as count (percentage). Rec. is the number recovered from raw failures and the percentage point gain. The Qwen rows form an evaluator only MASM ablation and were not included in the GPT 5.5 human correction study.}
\label{tab:main-results}
\scriptsize
\setlength{\tabcolsep}{5pt}
\renewcommand{\arraystretch}{1.08}
\begin{tabular}{lcrrr}
\toprule
Model and support & Steps & Raw $n$ (\%) & Corr. $n$ (\%) & Rec. $n$ (pp) \\
\midrule
Qwen3.5 9B, untrained & 25 & 23 (13.90) & -- & -- \\
Qwen3.5 9B, untrained + MASM & 25 & 31 (18.80) & -- & -- \\
GPT 5.5 + MASM & 15 & 49 (29.70) & 58 (35.15) & 9 (+5.45) \\
GPT 5.5 + Guide Text + MASM & 15 & 49 (29.70) & 61 (36.97) & 12 (+7.27) \\
GPT 5.5 + MASM & 25 & 43 (26.06) & 57 (34.55) & 14 (+8.49) \\
GPT 5.5 + Guide Text + MASM & 25 & 51 (30.91) & \textbf{63 (38.18)} & 12 (+7.27) \\
\bottomrule
\end{tabular}
\end{table*}

Across the four GPT 5.5 settings, manual review recovers 47 successful outcomes, an average of 11.75 tasks or 7.12 percentage points per setting. Individual corrections range from 9 tasks (5.45 percentage points) to 14 tasks (8.49 percentage points). The recovered cases are not marginal changes to a large score: they account for between 15.5\% and 24.6\% of the corrected successes in each setting. Common causes include semantically correct answers rejected by strict formatting, successful state changes missed by the final rule, and valid results followed by an incompatible stop or answer action. Raw and corrected results should therefore be reported together.

\subsection{Effects of MASM}

For the untrained Qwen3.5 9B model, MASM raises the evaluator score from 23 of 165 tasks (13.90\%) to 31 (18.80\%) under the same 25 step budget. This gain of eight tasks (4.85 percentage points) comes from changing the prompt and state representation rather than model weights. The action analysis asks whether the intended page transition occurred, and the memory update carries verified facts, obstacles, and next steps into the following action.

The same records also expose a limitation. Memory is a model generated state summary, not a verifier. Failed runs sometimes claim high completion despite a missing click, unsaved form, or absent final answer. MASM is therefore most useful when its structured record is checked against environment evidence, as in our review protocol.

\subsection{Effects of Guide Text}

Guide Text has a positive but nonuniform effect in both matched GPT 5.5 comparisons. At 15 steps, corrected success increases from 58 of 165 tasks (35.15\%) to 61 (36.97\%). Eight tasks (4.85\%) change from failure to success, while five (3.03\%) regress. At 25 steps, corrected success increases from 57 tasks (34.55\%) to 63 (38.18\%), with twenty Guide Text only successes (12.12\%) and fourteen MASM only successes (8.49\%). The exact paired tests are not significant at conventional thresholds, but the task flips remain informative because they show where procedural guidance changes behavior. Guide coverage provides a second check on this interpretation. Among the 143 tasks with an available guide, corrected success rises from 53 (37.06\%) to 56 (39.16\%), with seven gains and four regressions. The 22 tasks without a guide remain at 5 successes (22.73\%) in both conditions, with one gain offset by one regression. The overall improvement is therefore concentrated in tasks for which procedural evidence is actually present rather than in a general prompt change.

\begin{table}[H]
\centering
\caption{Corrected GPT 5.5 outcomes by Guide Text availability at 15 steps. Gain and loss count paired task flips relative to MASM only.}
\label{tab:guide-coverage}
\scriptsize
\setlength{\tabcolsep}{4pt}
\begin{tabular}{lrrrr}
\toprule
Subset & Tasks & MASM & Guide + MASM & Gain/loss \\
\midrule
All tasks & 165 & 58 (35.15\%) & 61 (36.97\%) & 8 / 5 \\
Guide available & 143 & 53 (37.06\%) & 56 (39.16\%) & 7 / 4 \\
No guide & 22 & 5 (22.73\%) & 5 (22.73\%) & 1 / 1 \\
\bottomrule
\end{tabular}
\end{table}

\begin{table}[H]
\centering
\caption{Matched GPT 5.5 comparison with and without Guide Text.}
\label{tab:guide-comparison}
\scriptsize
\setlength{\tabcolsep}{4pt}
\begin{tabular}{lrr}
\toprule
Measure & 15 steps & 25 steps \\
\midrule
Corrected MASM success & 58 (35.15\%) & 57 (34.55\%) \\
Corrected Guide Text + MASM success & 61 (36.97\%) & 63 (38.18\%) \\
Guide Text only & 8 (4.85\%) & 20 (12.12\%) \\
MASM only & 5 (3.03\%) & 14 (8.49\%) \\
Net change & +3 (+1.82 pp) & +6 (+3.64 pp) \\
Exact paired test & $p=.549$ & $p=.392$ \\
\bottomrule
\end{tabular}
\end{table}

At 25 steps, action tasks gain five successes and mixed tasks gain three, while retrieval tasks lose two (Table~\ref{tab:guide-types}). Shopping administration contributes four net successes, Reddit two, and GitLab and shopping one each; map tasks lose two. Guide Text is most helpful when it supplies an exact route, search phrase, or intermediate constraint. It can be harmful when an incomplete procedure anchors the model to the wrong page or omits the final confirmation.

\begin{table}[H]
\centering
\caption{Corrected GPT 5.5 success by task type at 25 steps.}
\label{tab:guide-types}
\scriptsize
\setlength{\tabcolsep}{5pt}
\begin{tabular}{lrrr}
\toprule
Task type & MASM & Guide Text + MASM & Net \\
\midrule
Action & 11 (6.67\%) & 16 (9.70\%) & +5 \\
Mixed & 26 (15.76\%) & 29 (17.58\%) & +3 \\
Retrieval & 20 (12.12\%) & 18 (10.91\%) & $-2$ \\
\bottomrule
\end{tabular}
\end{table}

The site breakdown in Table~\ref{tab:guide-sites} reinforces the task type result. Gains are concentrated in interfaces with relatively stable procedural routes: shopping administration gains four tasks, Reddit gains two, and GitLab and shopping gain one each. Map tasks lose two tasks, where a partial place or route hint can anchor the run before the page supplies enough evidence.

\begin{table}[H]
\centering
\caption{Net corrected Guide Text effect by site at 25 steps.}
\label{tab:guide-sites}
\scriptsize
\setlength{\tabcolsep}{6pt}
\begin{tabular}{lr}
\toprule
Site & Net change \\
\midrule
Shopping administration & +4 (+2.42 pp) \\
Reddit & +2 (+1.21 pp) \\
GitLab & +1 (+0.61 pp) \\
Shopping & +1 (+0.61 pp) \\
Map & $-2$ ($-1.21$ pp) \\
\bottomrule
\end{tabular}
\end{table}

\subsection{Task-Level Gains, Regressions, and Efficiency}

Aggregate success hides how strongly the intervention rearranges the solved task set. In the 25 step comparison, 34 of 165 tasks (20.61\%) flip outcome even though the net gain is only six. Representative gains include opening the correct best seller report, preserving direction and travel modes in a multi segment route, using a productive product query, and reaching the correct member management or forum subscription workflow. Regressions include following an incomplete nearby place hint, entering an irrelevant token page, losing the target forum context, and setting report dates without running the report. Appendix~\ref{app:task-flips} lists representative cases.

The intervention also changes efficiency. On the twenty tasks solved only with Guide Text, the mean action count falls from 15.0 to 8.9, the median from 14 to 7, and the number of runs reaching the 25 action limit from seven to one. This pattern is stronger evidence than the net score alone: useful procedural evidence can remove broad exploration, but a poor guide can move the agent efficiently toward the wrong state.

\subsection{Failure Modes and Completion Before Stopping}

We annotate the 102 failed trajectories from the strongest audited condition, GPT 5.5 with Guide Text and MASM at 25 steps. Table~\ref{tab:gpt-failures} reports the primary cause. Scroll oscillation is the largest category with 32 failures (31.4\%). Unfinished exploration and premature answering each account for 13 (12.7\%). Invalid action formats account for 9 (8.8\%), and incomplete form or edit workflows for 8 (7.8\%). Together, these categories describe different repair targets: loop detection, staged information gathering, completion checks, action validation, and explicit save or submit verification.

\begin{table}[t]
\centering
\caption{Primary failure modes for 102 failed GPT 5.5 trajectories at 25 steps with Guide Text and MASM.}
\label{tab:gpt-failures}
\scriptsize
\setlength{\tabcolsep}{4pt}
\begin{tabular}{lrr}
\toprule
Failure mode & $n$ & Share \\
\midrule
Scroll oscillation & 32 & 31.4\% \\
Exploration unfinished & 13 & 12.7\% \\
Premature answer & 13 & 12.7\% \\
Invalid action format & 9 & 8.8\% \\
Incomplete form or edit & 8 & 7.8\% \\
Exact action repetition & 7 & 6.9\% \\
Wrong value or entity & 6 & 5.9\% \\
Click without state change & 4 & 3.9\% \\
Search repetition & 3 & 2.9\% \\
Unsupported guess & 3 & 2.9\% \\
Navigation loop & 2 & 2.0\% \\
Other execution error & 1 & 1.0\% \\
Remaining defined modes & 1 & 1.0\% \\
\midrule
Total & 102 & 100\% \\
\bottomrule
\end{tabular}
\end{table}

A separate review flag records whether the requested state was reached before stopping. This distinction is necessary because a run can complete the browser operation and still fail through an invalid final action, an evaluator mismatch, or continued interaction that loses the correct state. Conversely, a 100\% statement in memory is not accepted as completion without page or task specific evidence. This evidence rule prevents optimistic self reports from being counted as recovered successes.

\subsection{Step-Level Progress and Trajectory Analysis}

Structured progress is extractable from memory for 91 of 165 GPT 5.5 tasks (55.2\%). Among all 102 failed runs, 18 (17.6\%) claim at least 80\% completion and 7 (6.9\%) claim 100\%. Human grounding substantially narrows these candidates. Among 40 analyzable failures that exhaust the action budget, the mean maximum verified progress is 56.2\%; eight reach at least 80\%, and none reaches 100\%.

\begin{table}[H]
\centering
\caption{GPT 5.5 progress evidence for the 25 step Guide Text and MASM condition.}
\label{tab:gpt-progress}
\scriptsize
\setlength{\tabcolsep}{4pt}
\begin{tabular}{lr}
\toprule
Measure & Value \\
\midrule
Structured progress extractable & 91/165 (55.2\%) \\
Failed runs reporting $\geq80\%$ & 18/102 (17.6\%) \\
Failed runs reporting $100\%$ & 7/102 (6.9\%) \\
Analyzable action limit failures & 40 \\
Mean maximum verified progress & 56.2\% \\
Action limit failures at $\geq80\%$ & 8/40 (20.0\%) \\
Action limit failures at $100\%$ & 0/40 (0.0\%) \\
\bottomrule
\end{tabular}
\end{table}

The first two actions yield the largest mean reported increments, 23.6 and 19.0 percentage points. The next four contribute 9.9, 8.6, 7.4, and 4.2 points (Table~\ref{tab:progress-increments}). The steep early increase is consistent with navigation and initial information gathering, while later actions are disproportionately associated with verification, recovery, repeated exploration, and stopping. Reported progress is useful for locating these transitions, but the gap between seven failed runs claiming 100\% and zero verified completions among the 40 action limit failures shows why the signal cannot stand alone.

\begin{table}[H]
\centering
\caption{Mean reported GPT 5.5 progress increments over the first six actions.}
\label{tab:progress-increments}
\scriptsize
\setlength{\tabcolsep}{4pt}
\begin{tabular}{lrrrrrr}
\toprule
 & Step 1 & Step 2 & Step 3 & Step 4 & Step 5 & Step 6 \\
\midrule
Increment (pp) & 23.6 & 19.0 & 9.9 & 8.6 & 7.4 & 4.2 \\
\bottomrule
\end{tabular}
\end{table}

This pattern explains why a longer budget does not automatically improve success: it creates more opportunity to recover, but also more opportunity to loop or leave a correct state. A trajectory aware monitor should therefore treat a plateau after rapid early progress as a signal to verify completion or change strategy, not simply spend the remaining budget.

\subsection{Implications for Reliable Web Agent Evaluation}

Three practices follow from the audit. First, benchmark reports should retain the original score but place correction rules, corrected outcomes, and recovered counts beside it. Second, prompt interventions should be compared by paired task gains, regressions, and action counts rather than only by net accuracy. Third, agent monitoring should combine final success with environment grounded progress, repeated action detection, and an explicit completion gate. These additions do not replace the benchmark score; they make clear what the score contains and where it fails.

\section{Limitations and Threats to Validity}

Each condition contains one rollout per task, so paired task flips describe the observed runs rather than a distribution over repeated trials. The 15 and 25 action budgets are not directly comparable; our strongest intervention claims therefore come from matched Guide Text comparisons at the same budget. The Qwen MASM ablation is reported with its original evaluator score and did not receive the same failure correction as the four GPT 5.5 settings. Human labels remain judgment based despite fixed evidence rules and second review for ambiguous cases. Structured progress is available for only 55.2\% of tasks and is used to locate evidence, not as ground truth. Guide Text is an observational intervention built from available procedural records rather than a randomized source: 22 tasks have no guide, and guide specificity varies by task. At least one action task shows possible state carryover from the preceding run, while another guide exposes an intermediate answer strongly enough to risk leakage. We retain these cases in the paired counts but flag them in the task level audit instead of interpreting every flip as a causal guide effect. Finally, API behavior, website state, and authentication can vary across reruns.

\section{Conclusion}

A final benchmark score is useful, but it does not fully describe either task completion or agent behavior. Across four GPT 5.5 settings, human review recovers 47 missed successes, corresponding to 5.45 to 8.49 percentage points per setting. The trajectory audit assigns 102 failed runs to reproducible causes and shows that high self reported progress often lacks sufficient environment evidence. MASM improves an untrained Qwen3.5 9B evaluator score, while Guide Text changes both the success rate and the identity of solved tasks for GPT 5.5. Reliable web agent evaluation should therefore report original and corrected outcomes together and preserve enough trajectory evidence to explain failures, progress, and stopping behavior.

\section{Ethical Consideration}

The retained study is evaluation only. GPT 5.5 outputs are used to evaluate the originating API model and to conduct human analysis of its recorded behavior. No GPT generated output or trajectory is used to train, fine tune, distill, or otherwise improve another model in the experiments reported in this paper. The records are used solely for this academic evaluation and are not used to develop, train, evaluate, or deploy a commercial product or service.

% References are excluded from the 9-page main-text limit.
\bibliographystyle{plainnat}
\bibliography{references}

\clearpage
\appendix
\setlength{\textfloatsep}{8pt plus 2pt minus 2pt}
\setlength{\floatsep}{8pt plus 2pt minus 2pt}
\setlength{\intextsep}{8pt plus 2pt minus 2pt}

\section{Complete 165-Task Outcome Matrix}
\label{app:matrix}

The complete matrix is organized by the common WebArena Lite task index. The six retained conditions store the original evaluator outcome, while the four GPT 5.5 conditions additionally store the human corrected outcome and correction rule. Failed audited runs include the first consequential error step, primary failure mode, maximum verified progress, and completion step when present. Table~\ref{tab:main-results} is the direct aggregate of this matrix.

\section{Human Review Guidelines and Correction Rules}
\label{app:review-rules}

A raw failure is corrected only when the requested result is supported by the recorded environment or final answer. Accepted corrections include semantically equivalent answer formats, harmless extra text, strict ordering differences when order is not requested, and successful state changes missed by a final string rule. We do not correct unsupported intentions, partially completed forms, inferred answers without page evidence, or a memory statement that conflicts with the environment.

Evidence is checked in the following order: task specific environment state, before and after screenshots, execution result, final answer, and model Memory or Analysis. The reviewer records the evidence and one correction code. Ambiguous cases are sent to a second reviewer rather than resolved from model self report.

\section{Definitions of the Fine-Grained Failure Modes}
\label{app:failure-modes}

\begin{table*}[!htbp]
\centering
\scriptsize
\setlength{\tabcolsep}{3pt}
\renewcommand{\arraystretch}{1.08}
\begin{tabular}{p{3.2cm}p{9.4cm}}
\toprule
Failure mode & Operational definition \\
\midrule
Scroll oscillation & Repeated up and down scrolling without extracting new evidence or changing strategy. \\
Exploration unfinished & The run exhausts its budget before reaching the page or state needed for the task. \\
Premature answer & A final answer is emitted before required evidence or action confirmation is obtained. \\
Wrong value or entity & The agent selects a similar but incorrect item, person, date, count, or destination. \\
Click without state change & A click leaves the relevant URL, page, and task state unchanged and is not repaired. \\
Exact action repetition & The same executable action is repeated in the same state. \\
Search repetition & Queries are repeated or reformulated without convergence on verifiable evidence. \\
Incomplete form or edit & Required input, save, submit, or confirmation steps are missing. \\
Invalid action format & The tool call cannot be executed because its action name, element, or parameter is invalid. \\
Navigation loop & The agent repeatedly moves between pages or backtracks without preserving progress. \\
Other execution error & A valid high level plan fails because of an uncategorized execution or interface error. \\
Unsupported guess & The answer is not grounded in the observed page or environment state. \\
Strict answer mismatch & The task is complete, but a semantically valid answer is rejected by a narrow output rule. \\
Missing answer field & The answer omits one of several values explicitly required by the instruction. \\
Wrong page or site & The agent acts in an incorrect project, forum, report, product, or destination context. \\
Completion before stopping & Required state is reached, but the run continues, corrupts the state, or fails to emit a valid stop. \\
\bottomrule
\end{tabular}
\end{table*}

The complete count distribution for the 102 failed GPT 5.5 runs is reported in Table~\ref{tab:gpt-failures}. The primary label records the first cause that materially reduces the chance of completion; later symptoms remain in the reviewer note.

\section{Per-Task Guide Text Gains and Regressions}
\label{app:task-flips}

Table~\ref{tab:guide-examples} lists representative flips behind the aggregate task type and site results in Tables~\ref{tab:guide-types} and~\ref{tab:guide-sites}. Counts are paired by task identifier and use human corrected outcomes.

\begin{table*}[!htbp]
\centering
\scriptsize
\setlength{\tabcolsep}{4pt}
\begin{tabular}{rll}
\toprule
Task index & Guide Text effect & Observed mechanism \\
\midrule
0 & gain: January top three best sellers & Direct route to the report and correct date interval. \\
18 & gain: two segment travel time & Preserves segment direction and travel modes. \\
29 & gain: product search for bruxism & Supplies a useful query and explicit completion condition. \\
130 & gain: add two GitLab maintainers & Provides the correct members page workflow. \\
132 & gain: subscribe to a trending forum thread & Preserves forum context and verifies subscription. \\
12 & regression: nearest cafe to Hunt Library & Incomplete guide anchors the model to an unsupported cafe. \\
63 & regression: RSS feed token & Redirects the model to an irrelevant personal access token page. \\
134 & regression: post a console question & Generic submit guidance loses the target forum context. \\
149 & regression: February sales report & Dates are set, but the report is not run and checked. \\
\bottomrule
\end{tabular}
\caption{Representative Guide Text gains and regressions from paired GPT 5.5 runs.}
\label{tab:guide-examples}
\end{table*}

For the twenty Guide Text only successes at 25 steps, mean actions fall from 15.0 to 8.9, median actions from 14 to 7, and action limit cases from seven to one. These efficiency values are computed on the paired flip set rather than on all tasks.

\section{Step-Level Progress Annotation Protocol}
\label{app:progress-protocol}

Annotators first decompose the instruction into atomic conditions that can be checked from the environment. At each step they mark the currently verified conditions and record the evidence source. Current progress may decrease when a correct state is overwritten. We separately retain attained progress, $\widehat{P}(t)=\max_{s\leq t}P(s)$, so later forgetting does not erase evidence that the task had already been complete. Overrun is the number of actions after the first verified completion, excluding one action needed to emit a final answer or stop.

Memory progress is used to select candidate steps, not to assign the final label. The seven failed GPT 5.5 runs with a 100\% memory report are therefore reviewed against screenshots, action results, and the task rubric. None of the 40 analyzable failures that exhaust the action budget reaches 100\% verified progress.

\section{Representative Audited Trajectories}

Task 35 asks for the driving time from Joe Biden's hometown to Bridgeport. The GPT 5.5 run recovers approximately three hours and twenty four minutes and reports full progress by step four. It then generates invalid final action formats before a correct exit is still recorded as an execution error. The browser reasoning succeeds, but stopping syntax and evaluation do not agree.

Task 149 illustrates the opposite pattern. The agent identifies the intended reporting period and enters the dates, but stops before running and checking the report. The final state is genuinely incomplete, so no correction is applied even though the memory reports substantial progress. These two cases show why outcome correction and trajectory diagnosis must use the same environment evidence.

\section{Review Interface and Analysis Tool}

The review tool aligns the retained evaluation conditions for each task on one page. It displays original and corrected outcomes, automatic failure summaries, action counts, before and after screenshots, action analysis, memory, and the final response. Reviewers record the verdict, failure mode, first error step, corrected answer, and notes. Filters expose evaluator disagreements, prompt specific failures, task types, and unreviewed cases. Using one interface for correction and trajectory adjudication keeps the evidence visible and the criteria consistent.

\section{Evaluation and Environment Details}

All runs use the 165 WebArena Lite task identifiers. Evaluations are executed after website reset and authentication state recreation. GPT 5.5 is called directly under four prompt and budget conditions: MASM at 15 and 25 steps, and Guide Text with MASM at 15 and 25 steps. The untrained Qwen3.5 9B model is evaluated with and without MASM at 25 steps. Task order, reset procedure, prompts, action records, screenshots, evaluator outputs, and human correction records are retained with the study artifact. No recorded GPT output is used as training data in the experiments reported in this paper.

\end{document}